%% file: main.tex
\documentclass[a4paper,UKenglish,cleveref, autoref, thm-restate]{lipics-v2021}
\nolinenumbers
\hideLIPIcs
\usepackage{graphicx} 
\usepackage{lipsum}
\usepackage{url}
\usepackage{amsfonts}
\usepackage{amsmath}
\usepackage{amssymb}
\usepackage{cleveref}
\usepackage{nicefrac}
\usepackage{subcaption}
\usepackage{wrapfig}
\usepackage{booktabs}
\usepackage{hyperref}
\usepackage{placeins}

\title{Hybrid Reinforcement Learning and Search\\for Flight Trajectory Planning}
\date{March 2025}
\titlerunning{Integrating RL with Search \& Optimization strategies}
\author{Alberto Luise}{University of Bologna, Italy}{alberto.luise2@unibo.it}{}{}

\author{Michele Lombardi}{University of Bologna, Italy}{michele.lombardi2@unibo.it}{}{}
\authorrunning{A. Luise et al.}

\Copyright{Alberto Luise, Michele Lombardi, Alistair Nottle, and Florent Teichteil Koenigsbuch}

\ccsdesc{Computing methodologies~Artificial intelligence}

\begin{document}

\maketitle    
\begin{abstract}
    This paper explores the combination of Reinforcement Learning (RL) and search-based path planners to speed up the optimization of flight paths for airliners, where in case of emergency a fast route re-calculation can be crucial.
    The fundamental idea is to train an RL Agent to pre-compute near-optimal paths based on location and atmospheric data and use those at runtime to constrain the underlying path planning solver and find a solution within a certain distance from the initial guess. The approach effectively reduces the size of the solver's search space, significantly speeding up route optimization. Although global optimality is not guaranteed, empirical results conducted with Airbus aircraft's performance models show that fuel consumption  remains nearly identical to that of an unconstrained solver, with deviations typically within 1\%. At the same time, computation speed can be improved by up to 50\% as compared to using a
    conventional solver alone.
    This paper discusses the theoretical framework, the different implementation strategies, the adopted testing procedures, the obtained results and finally further possible developments and future perspectives. 

\keywords{Reinforcement Learning  \and AI \and Hybrid methods \and Constraint programming}

\end{abstract}




\section{Introduction}

The optimal trajectory for an airliner to go from point A to point B is often not a straight line, not only due to the Earth's curvature but also due to weather conditions, wind, temperature and other atmospheric factors. Furthermore, fully simulating an aircraft is not an easy task, as they are very complex machines: computing the transition states and costs from one waypoint to another in a search graph requires to call an external performance model simulator which is very CPU time consuming. Therefore, standard flight planning algorithms rely on a number of approximations ranging from simplified performance model to decoupled lateral/longitudinal search. In order to allow for practical use of more accurate and non-decoupled flight planning models as the ones developed by Airbus in the \textit{scikit-decide} library \cite{skdecide}, more advanced techniques like hybridization of machine learning and model-based planning are necessary.
While computation speed might not be a problem when computing an airport-to-airport route in advance, it could become crucial depending on the application. In fact, in case of \textbf{Flight Diversion}, an automated planner could be used on-line to re-compute multiple trajectory options to the nearest airports in case a passenger has a critical health condition and has to be taken to a hospital as soon as possible. In these circumstances a slow planner, accurate as it may be, is not adequate.

The solution proposed in this paper is to build over an existing route construction approach, based on combinatorial search and light constraint propagation via Forward Checking.
Specifically, inspired by the UNIFY framework \cite{unify}, a coarse route construction stage based on Reinforcement Learning (RL) is introduced, and used to constrain the existing combinatorial approach to search in the vicinity of the RL routes.
The resulting hybrid method can reduce the computation time by up to 50\%, thanks to the low inference cost of RL, with a negligible or moderate loss in terms of solution quality.

Additionally, while the focus of this paper is on on a deterministic version of the problem, the method could be naturally extended to account for uncertainty.

\input{problem}

\input{model}
\input{Experiments}

\section{Conclusions}

This paper has explored the integration of Reinforcement Learning into the planning strategies used for flight planning support, using realistic Airbus aircraft performance models and weather data. The primary objective was to enhance the efficiency of aircraft trajectories computations, leveraging the strengths of RL to complement traditional path planning techniques.
A novel hybrid solution was proposed, combining RL and search techniques to expedite the process of route optimization: the agent pre-computes an approximate trajectory to be used as a starting point, reducing the search space for the solver and thus accelerating computation.
The integration of RL proved effective in maintaining optimal fuel consumption while significantly reducing computation time, especially in complex scenarios involving larger graph sizes. It should be noted, however, that reducing exceedingly the search will yield sub-optimal solutions.
Empirical results demonstrated that the hybrid model can reduce computation time by around 40\% compared to the standalone solver on a standard setup, with a width value immediately below half the graph's size and without sacrificing the quality of the solution. This improvement and increase up to 50\% for particularly large graph searches. As efficiency is crucial in emergency scenarios where rapid decision-making is essential, such as Flight Diversion, the hybrid model shows promising results.

\subsection{Future developments}

This work has been designed in a way that accommodates the inclusion of future works on \textbf{uncertainty}, which could be handled well by a Reinforcement Learning agent. Every piece of data regarding weather has been treated as deterministic, which is fine for a simulation but unrealistic for a real world application: sensors can be imprecise, predictions can be inaccurate. Furthermore, RL could also be used to predict and avoid extreme weather conditions, such as incoming storms, by analyzing surrounding indicators (to be defined with the help of an expert). Overall, creating an hybrid model with the inclusion of a RL agent has shown to provide huge benefits with little trade-off, provided that parameters are chosen accordingly, showing potential for both real world application and future research. 

\newpage

\bibliographystyle{plainurl}
\bibliography{ref}

\input{Appendix}

\end{document}

%% file: problem.tex
\section{Problem Definition}
This paper considers the problem of planning a route for an aircraft from an origin to a destination, accounting for earth geometry, weather, and fuel consumption through a deterministic model.
Such a trajectory is generally specified via a sequence of waypoints, starting with the origin and ending with the destination.
Navigation between subsequent waypoints is assumed to follow a great circle trajectory, with no adjustment made for adverse weather or obstacles.
In this paper, reducing fuel consumption is the main objective, since it accounts for flight time, but also for potentially difficult encountered weather conditions.

Formally, the state of an aircraft can be modeled as a pair $(x, s)$, where $x \in \mathbb{R}^3$ represents the aircraft location in a three-dimensional space and the vector $s$ captures all remaining state components, such as mass, acceleration, fuel level, etc.
Weather conditions include the wind speed, wind direction, and temperature and are modeled through a tensor field $\xi(x)$, associating relevant information to each location.
The cost of flying between two locations $x'$ and $x''$ along a great circle trajectory is specified via a function $f_\xi(x', x'')$, which is evaluated by means of a simulator.
An aircraft route consists of a sequence $\tau = \{x_k\}_{k=1}^n$, where $x_1$ represents the origin and $x_n$ the destination.
The cost of traversing the route is given by: 
\begin{equation}
    f_\xi(\tau) = \sum_{k=1}^{n-1} f_\xi(x_k, x_{k+1}).
\end{equation}
This work assumes the availability of an approximation algorithm capable of computing near-optimal routes; 
in this specific case such an algorithm consists in a planner that works by:
1) defining a discretization of the three-dimensional space between the origin and the destination;
2) building a directed-graph representation over the discretization, using forward-checking to account for feasibility of flight maneuvers;
3) computing the cost of moving from connected nodes in the graph via a linear approximation of the cost $f_\xi(x', x'')$, for reduced computational cost;
4) searching for an optimal path via A*, guided by a custom heuristic.
The algorithm is implemented in the \textit{scikit-decide} library, an AI framework for Reinforcement Learning, Automated Planning and Scheduling \cite{skdecide}. It has been co-developed at Airbus AI Research together with the ANITI research center and TUPLES project and provides multiple solvers capable of interacting with pre-existing or custom built environments. Specifically, while the project is still being developed and maintained, version 1.0.2 has been used for the purposes of this paper.

The space discretization consists of a three-dimensional box, whose number of nodes along each axis can be configured.
A finer discretization results in a better approximation of the cost, a better representation of the continuous nature of aircraft travel, and therefore in remarkably better route costs.
Unfortunately, a linear increase of the number of nodes on every axis leads to a cubic increase in the total number of nodes.
Together with the complex nature of the problem cost, this results in run-times that are not suitable for online scenarios, where flight diversions need to be quickly computed.
The main goal of this contribution is that of \emph{reducing the time needed to compute a route}.
Additionally, this work aims at paving the way for a solution approach capable of accounting for uncertainty in the weather condition fields, which is currently treated as deterministic.

%% file: model.tex
\section{Hybrid model}

The structure of a solution for the considered problem (i.e. a sequence of locations) suggests that significant run-time reductions could be reached via a hierarchical approach \cite{ijcai2016p429,ijcai2019p875}:
for example, one could think of planning a route with a coarse discretization, then connecting the resulting waypoints with routes built using a finer discretizaton.
Unfortunately, using a coarse discretization makes the linear cost approximation used by the planner very inaccurate.
Additionally, forcing the route to traverse the waypoints built with a coarse discretization can be overly restrictive, especially if such waypoints are poorly chosen.
Finally, such a hierarchical approach could not be easily extended to deal with weather uncertainty.

The proposed idea is to build a ``soft'' hierarchical approach, where the problem is first tackled by a faster, less accurate approach (traversing a continuous space instead of being constrained by a waypoint graph) and afterwards this information is used by a more thorough planner to solve the task from the beginning but with a clear guideline for the optimal solution.
The high-level route needs to be calculated using an algorithm with a very fast run time, capable of dealing with non-linear environments and possibly ready to deal with uncertainty; a perfect fit for this description is \textbf{Reinforcement Learning}.
The low-level route can then be calculated by the scikit-decide planner using $A^*$, but with an additional constraint forcing it to explore only the graph within a fixed distance from the previously computed trajectory, composed of a few intermediate points in space.
This results in a significant reduction of the search space and of the computation time.

\subsection{The Reinforcement Learning Agent}
\label{sec:agent}

As previously stated, the task of the RL agent is to build a coarse trajectory, including only a small pre-decided number $n$ of waypoints.
In particular, the agent is formally a function taking as input a destination $x_n$ and the current aircraft location $x$, and providing as output an action.
The function is actually the composition of a feature extraction module and a Neural Network (NN), i.e. $a_\theta(g(x, x_n))$.
Starting from the current location and the destination, the feature extractor provides to the NN:
1) the difference vector between the destination and the current location, rotated by an angle $\phi$ in order to make every instance of the problem appear as a straight trip upwards and deprived of the altitude component;
2) all information from the weather field.
Formally, we have that:
\begin{equation}
    g : \begin{pmatrix}x \\ x_n\end{pmatrix}
    \mapsto
    \begin{pmatrix}
    R_\phi I_2 (x_n - x) \\
    \xi(x)
    \end{pmatrix}
\end{equation}
where:
\begin{equation}
    R_\phi = \begin{pmatrix}
        \cos(\phi) & - \sin(\phi) \\
        \sin(\phi) & \cos(\phi)
    \end{pmatrix}
    \qquad
    I_2 = \begin{pmatrix}
        1 & 0 & 0 \\
        0 & 1 & 0 \\
        0 & 0 & 0
    \end{pmatrix}
\end{equation}
%
The $I_2$ matrix is used to filter out the altitude component from the locations.
The RL agent does not handle altitude directly, for multiple reasons:
\begin{itemize}
    \item The impact of altitude on the horizontal projection of the flight plan in term of lateral deviations is typically lower than the width of the corridor used by the low-level flight planner. Therefore, taking into account altitude in the RL agent would not change much the resulting coarse trajectory, thereby the corridor used by the flight planner.
    \item Considering altitude would increase the number of dimensions of both the state and the action space for the Agent, further increasing the difficulty of the learning problem.
    \item The search graph used by the planner is often quite narrow along the vertical component (due to altitude control), and therefore plays a minor role in the research cost (and even if it had a major impact, vertical components marginally impact the horizontal profile).
    \item Designing an agent that is insensitive to altitude allows one to avoid retraining in case an altitude profile is imposed by flight regulators, since these only require adjustments to the $A^*$ planner.
\end{itemize}
%
Based on this information, the NN $a_\theta$ computes an output action, consisting of a two dimensional vector in $[-1, 1]^2$, which is then scaled up and rotated to obtain a movement vector on the two dimensional plane.
This is used to compute the next aircraft location as:
\begin{equation}
        x_{k+1} = x_k +
        \begin{pmatrix}
            R_\phi^{-1} a_\theta(x, x_n) \frac{\|x_n - x_1\|_{2}}{n} \\
            \Delta h_k
        \end{pmatrix}
\end{equation}
where $x_1$ is the route origin and $\Delta h_k$ is the difference in altitude for the current step, according to the altitude profile.
The maximum distance the plane will be allowed to travel in a direction each step is the total length of the trip divided by the number of sub-steps calculated in the path.
At training time, all movement vectors are generated according to the agent output, meaning that the destination might not be reached exactly.
At inference time, the actual destination is always used as the last waypoint.



Give that the planning problem cost consists of the fuel consumption, one may think of using its (negated) value as the reward for the RL agent.
Doing this would however result in naïve (and physically meaningless) policies, since fuel consumption can be minimized by just keeping the aircraft stationary.
The Agent needs information guiding it towards the destination, and for this reason the reward consists of two components. The first one is \textbf{the component of the movement vector which contributes to reaching the destination}, obtained by projecting it onto the distance vector from start to finish, normalized by the sub-step length. It can be defined as 
\begin{equation}
    \label{eq:value}
        V_{k} = \frac{\| (\Delta x_k \cdot D_k) / (D_k \cdot D_k) \times D_k \|_{2}}
        {\| \Delta x_k \|_{2}} - \lambda
\end{equation}
where $\Delta x_k = x_{k+1} - x_{k}$, $D_k$ is the distance vector between $x_k$ and the destination.
For this computation, the altitude component is removed from both vectors.
Finally $\lambda$ is a small regularization parameter preventing $V_k$ from reaching 1.
The second component is actually the fuel consumption.
In detail, the reward function is $- (1 - V_{k})^{\gamma} \cdot F_k$ where $F_k$ is the consumed fuel $f_\xi(x_{k}, x_{k+1})$
and $\gamma$ is an hyperparameter regulating the importance given to both components of the equation. The $\lambda$ term inside $V_k$ prevents multiplication by zero.

The training method used for the agent is part of the Proximal Policy Optimization family, containing algorithms that learn a policy without having an explicit model of their environment.
Specifically it's an adaptation of \textit{PPO-clip}, which in addition to the standard Actor-Critic approach\footnote{Actor-Critic approaches combine a policy (the Actor), which samples the actions, and a differentiable value function (the Critic) that evaluates the rewards and provides low-variance gradients \cite{a2c}.}
employs a clipping mechanism, which improves stability preventing overly large changes in a single step.
Its inner workings and formulas are detailed in \cite{schulman2017}.
The agent was trained over 16,000 random instances of the Flight Planning problem, obtained by sampling two points per trip (origin and destination) from the European sky, coarsely represented by a rectangular area between $71^{\circ}N, 35^{\circ}E$ (above Finland) and $34^{\circ}N, 10^{\circ}W$ (Canary Islands).
If the distance between the sampled points is lower than a threshold $t$, the pair is re-sampled.
In all our experiments, we train the agent to generate routes including five waypoints, hence $n = 5$. These include the origin and the destination.
%
\begin{wrapfigure}{r}{0.5\textwidth}
    \centering
    \includegraphics[width=0.85\linewidth]{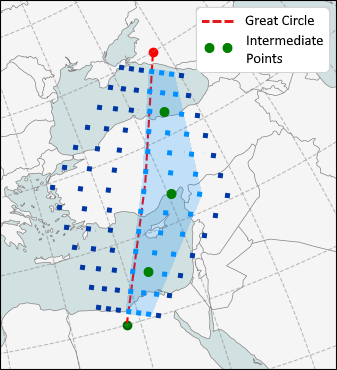}
    \captionof{figure}{Hybrid Model workflow: the blue points are the original graph, the RL route is in green, and the azure area is the constrained region.}
\end{wrapfigure}
Despite a single training routine taking up to 10 hours (making grid search especially difficult), at runtime the Agent, independently from the trip length, takes $\sim1.5s$ to calculate the approximate trajectory.

\subsection{The Complete Model}
\label{sec:hybrid_model}

Once the Agent's calculations are completed, the Flight Planning Domain is invoked and the graph for the trajectory search is built.
Such search graph is a three-dimensional matrix of nodes with a latitude, longitude and altitude component and with size $\{I, J, H\}$. Its first and last rows are composed of identical nodes, representing the starting point and the arrival point. 
The $A^*$ solver explores this network as a directed graph, where each node $n_{i,j,h}$ is only able to reach the ones in the successive row moving within adjacent values of $j$ and $h$, i.e. $n_{i+1, j', h'}$, where $j' \in \{j-1, j, j+1\}$ and $h' \in \{h-1, h, h+1\}$.

Information from the route computed by the RL agent is incorporated by marking some of the nodes as unreachable based on their distance from the approximate trajectory. Specifically, while $i$ represents the distance from the objective and $h$ is the altitude component only handled by the planner, the $j$ values will be used to define the optimality region acting as a threshold for reachable points (modifying accordingly the starting point of the algorithm in the first row of identical points).
The final search space is thus calculated as follows:
\begin{itemize}
    \item for each row, an intermediate point lying on the approximate trajectory of length $m$ is calculated as $\overline{x}_i = x_p + (x_{p+1} - x_p) (i \mod s)/s$, where  $s = I / m$ and $p = i / s$;
    \item after that, an interval $R =(\overline{j}_{min}; \overline{j}_{max})$ is defined, such that $\overline{j}_{min} = argmin_i(\overline{x}_i)$ and  $\overline{j}_{max} = argmax_i(\overline{x}_i)$;
    \item the region is then expanded or shrunk according to two parameters, $w_{min}$ and $w_{max}$: those represent the minimum and maximum allowed percentage of the original graph to be considered by the hybrid planner, and the margins of the interval $R$ are increased or decreased symmetrically until its length fits between the two parameters. 

    (if $\overline{j}_{max} - \overline{j}_{min} < w_{min}$, increase  $\overline{j}_{max}$ by 1 and decrease $\overline{j}_{min}$ by 1;
    
    \vspace{-0.1cm}
     if $\overline{j}_{max} - \overline{j}_{min} > w_{max}$ then decrease  $\overline{j}_{max}$ by 1 and increase $\overline{j}_{min}$ by 1.)
\end{itemize}

Note that our pipeline only shrinks the search space along the horizontal axis and not the vertical axis: this is because our RL Agent does not take into account multiple levels of altitude, and even if it is only meant to be a broad guideline for the underlying planner, it is not able to give any insight on the correct level of altitude for the plane. 

Once this procedure is complete, the \emph{sk-decide} planner is called to solve the instance on the newly formed graph, with the starting point of the search being $x_{0, (\overline{j}_{max} - \overline{j}_{min})/2, H/2}$.
By changing the values of $w_{min}$ and $w_{max}$, one can can control the extent of graph being searched. A sufficiently large region recovers the original planner behavior.

%% file: Experiments.tex
\section{Experiments}
\label{sec:tests}

All experiments were performed on a 8-Core, 3593 Mhz CPU, with 16 GB of RAM and a SSD with a sequential read speed of 3068 MB/s, operating inside an Anaconda environment on Windows 11. 

\subsection{Agent Training}

The RL training took place as described in Section \ref{sec:agent}, implemented using the Gymnasium library \cite{gymenv} for the environment and Stable-Baseline3 \cite{sb3} for the Agent. Grid search was employed to optimize the following hyperparameters:
    \begin{itemize}
        \item \texttt{Hidden Size}: the number of nodes in each hidden layer;
        
        \item \texttt{Clip Range}: bounds that limit how much the policy can change during training, represented by the letter $\epsilon$. It's typically set to a small value in $[0.1, 0.25]$;
        
        \item \texttt{Learning Rate}: used in gradient descent training;        
        
        \item \texttt{Reward Exponent}: represented by the letter $\gamma$ in the loss function, it determines how much importance is given to the direction taken by the plane or to the fuel consumed;
        
        \item \texttt{Extra End Reward}: flag boosting the last reward given to the Agent if the destination is reached (or hindering it in case of a missed flight). Given a threshold $T$ used to set a minimum distance  required from the precise destination to consider the approximate path as "valid", the added value to the final reward is calculated as follows:
        \begin{equation}
            r(P_k, T)_k =  \begin{cases}
                -\rho_1 * D(P_k, T) \;\;\;\;\;\; D(P_k, T) > T \\
                \rho_2 * (1 - D(P_k, T) ) \;\;\;\;\;\; D(P_k, T) < T
                \end{cases}
        \end{equation}        
    \end{itemize}
Where $\rho_1$ and $\rho_2$ are weights used to give more importance to the rewards to be delivered, in this case set respectively to 5 and 2. The optimal values found for each of these parameters are reported in \cref{tab:hyperparameters}.

\begin{table}[h]
    \centering
    \begin{tabular}{|ccccc|}
    \toprule
    Hidden Size & Clip Range & Learning Rate & Exponent & End Reward \\ \midrule
    64 & 0.2 & $5 \times 10^{-6}$ & 2.0 & True \\
    \bottomrule
    \end{tabular}
    \caption{Optimized hyperparameter values}
    \label{tab:hyperparameters}
\end{table}

\subsection{Test results}
\label{sec:results}

The main goal of this contribution is increasing computation speed without losing optimality.
We start by investigating whether the hybrid approach can compute routes with a fuel consumption similar to that of the original approach.
These results are not easy to compute, as the solver itself only returns a path to be traversed on the graph, without the cost values associated with it. Therefore, after each planning operation, a \textit{complete rollout} of the trajectory is necessary to compute the fuel consumption, and this implies a complete simulation of flight between each arc of the path. For this reason, extensive testing of extreme cases with more complicated and expensive routes was not feasible.

The trajectories computed for testing the approaches are selected at random depending on a \texttt{short / long} flag: taken a random starting point on Earth (chosen by selecting values between -90 and +90 for Latitude and -180 and 180 for Longitude) and a random direction, the destination point is selected along the resulting vector at 300-400km for \texttt{short} trips and 700-800km for \texttt{long} trajectories.

Each test flight can depend on a number of parameters:
\begin{itemize}
    \item \textbf{\texttt{Forward Points}}: Number of waypoints of a single trajectory. Modifying this parameters corresponds to increasing the density of the search graph (directly influences its $I$ component).
    \item \textbf{\texttt{Lateral Points}}: Number of possible straight trajectories composing the graph (directly influences its $J$ component).
    \item \textbf{\texttt{Graph Width}}: taking only values between \textit{"small"}, \textit{"medium"} and \textit{"large"}, it modifies how the graph is drawn and how far the points are from one another. A short graph will draw all points closer to the Great Circle trajectory, a large graph the opposite (see Figure \ref{fig:graphsize} for reference).
    \item \textbf{Trip Length}: the distance between origin and destination, defined as discussed above.
\end{itemize}

Before we can test how the different models perform in different settings, however, we must first study how the model can best reduce the search size of the algorithm without hindering the actual result. For this reason, before ultimating the pipeline, we made a sensitivity study on two random trajectory: a longer one of 1200km and a shorter one of 450km, with a graph of size (21,11,1). 

\begin{figure}[h]
    \centering
    \includegraphics[width=1\linewidth]{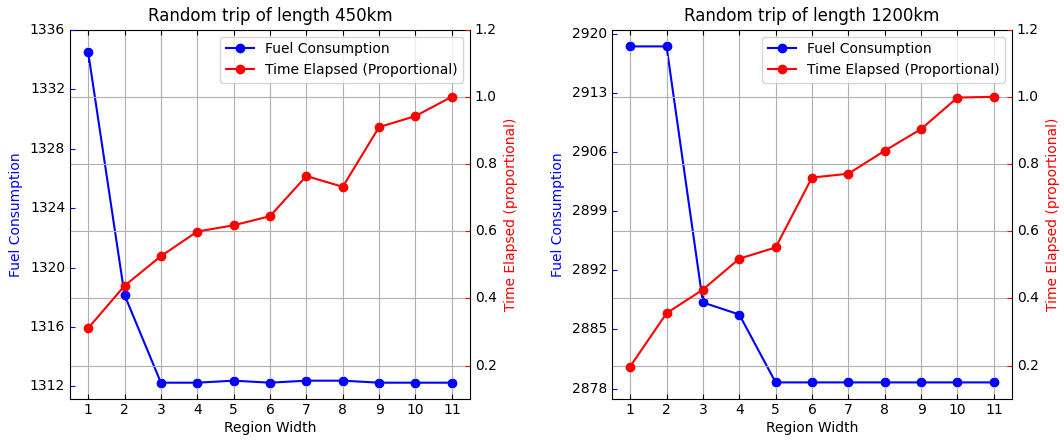}
    \caption{Analysis of computation speed relatively to changes in the width of the optimal region. While Fuel Consumption is displayed with its absolute values, on the left axis, on the right axis the Solving Time is shown as a percentage w.r.t. the value computed by the planner alone (which as can be seen in Figure \hyperref[fig:width_fuel]{\ref*{fig:width_fuel}.b} coincides with the Hybrid Model solution with maximum width), in order to best understand the contribution of the model.}
    \label{fig:width_fuel}
\end{figure}

 As can be seen in the comparison, the optimal value (in terms of fuel consumption) for $w$ is not consistent across various lengths and should therefore be chosen accordingly. However, it is also evident that a threshold after which the cost function remains optimal exists, in this case being an Optimal Region of width 5.
For this reason, and given the preceding analysis, a value for $w$ immediately below half the graph size is recommended to maintain optimality, but it can be lowered for shorter trips that require less profound exploration.

Therefore, instead of manually deciding the correct width parameter for every trip, a value for $w$ is decided at runtime based on the width of the graph and on the Agent's trajectory (as explained in Section \ref{sec:hybrid_model}) 

\begin{figure}[h]
    \centering
    \includegraphics[width=0.8\linewidth]{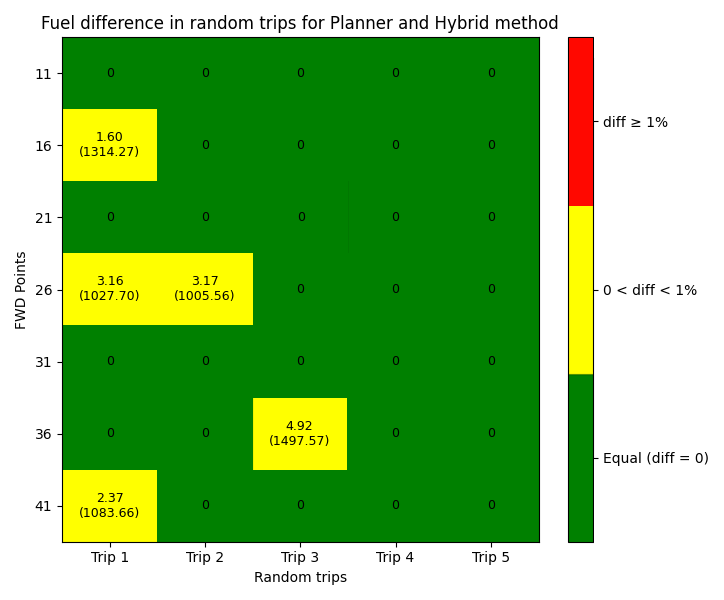}
    \caption{Table containing 5 random trips at each different FWD points value. Each cell represents the difference in fuel consumption between the standalone planner and the hybrid model, and when this quantity is different from zero the total value of fuel used for that trip is specified below in parentheses.}
    \label{fig:fuel_consumption}
\end{figure}

In Figure \ref{fig:fuel_consumption}, the results of 5 random trips for each of the previously analyzed FWD Points values are displayed; specifically, the difference between the results fo the two approaches. Note that not only 6 out of 7 calculations are exactly correct, but when a slight error is introduced this is always below a margin of 1\%.

\begin{figure}[h]
    \centering
    \includegraphics[width=1\linewidth]{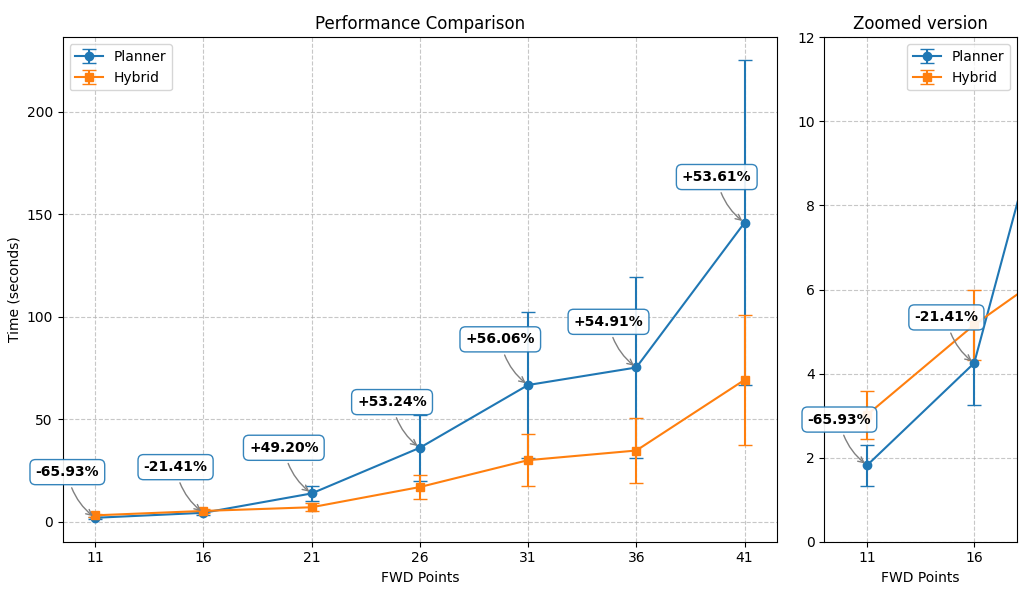}
    \caption{Analysis of computation speed relatively to changes in the density of the search graph, done modifying the number of FWD Points. The data shown is an average of 10 trips between random locations across the globe, on \texttt{short} distances.}
    \label{fig:speed_std}
\end{figure}

In Figure \ref{fig:speed_std} a simple speed comparison between the original approach and the hybrid model is shown, against the number of Forward (FWD) points. The plot representing the our approach's computation time also takes into account the time necessary to build the Agent's environment, which is not needed for the standalone planner.
While to an increase in FWD Points generally
corresponds an increase in computation time, this is not a fixed rule and can
present exceptions.
For example, an advantageous alignment of the specific network with the ground truth optimal trajectory or by the graph becoming large enough that correcting early mistakes has a broader impact on the results. In any case, for graphs longer than 26 FWD points (the default value for the planner is 41) the hybrid model sees a noticeable time consumption reduction, reaching up to a 50\% gain for 51 points.
Note also that these values are averages of trips of different length for which calculation time varies heavily, as can be noted by the Standard Deviation in both comparisons.

%
%

\subsection{Graph variations}

\begin{figure}[h]
    \centering
    \includegraphics[width=1\linewidth]{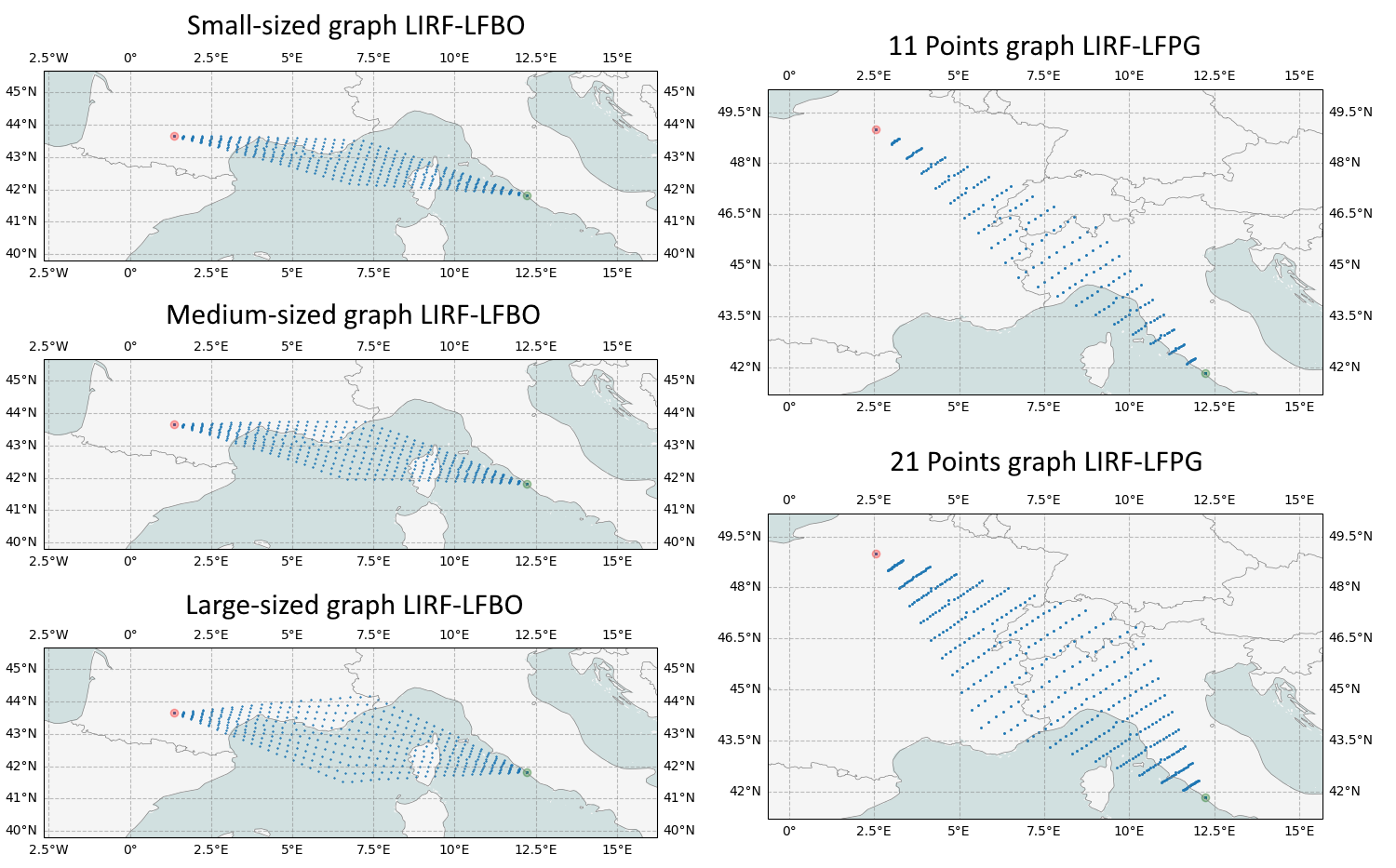}
    \caption{Visualization of different graph sizes for the same trips. On the left side, the LIRF-LFBO graph retain the same number of nodes but the \texttt{Graph Width} parameter changes how far they are from each other; on the right side, the LIRF-LFPG graph retains the same width but increases the number of available points to be explored.}
    \label{fig:graphsize}
\end{figure}

The overall shape and density of the graph, aside from the n. of FWD Points which directly influences the complexity of the problem, can be influenced by two parameters: \texttt{LAT Points} and \texttt{Graph width}. The first one dictates how many parallel lines compose the network, the second how far they are from each other (see Figure \ref{fig:graphsize} for reference). All previous examples were calculated at a \texttt{normal} graph width, with 11 \texttt{LAT Points}; a larger graph usually means longer distances (and therefore a longer simulation), while an increased number of parallel routes results in more choices to be evaluated by the planner (and hopefully more room for improvement).

\FloatBarrier

\begin{figure}[h]
    \centering
    \begin{subfigure}{0.49\linewidth}
        \centering
        \includegraphics[width=\linewidth]{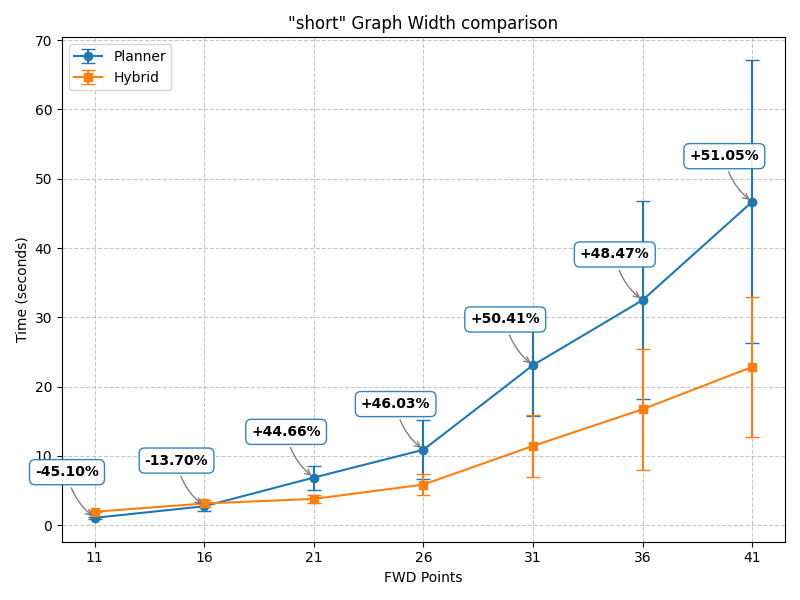}
        \caption{Comparison on thin graphs; constant increase up to +50\%.}
        \label{fig:graph_var_a}
    \end{subfigure}
    \hfill
    \begin{subfigure}{0.49\linewidth}
        \centering
        \includegraphics[width=\linewidth]{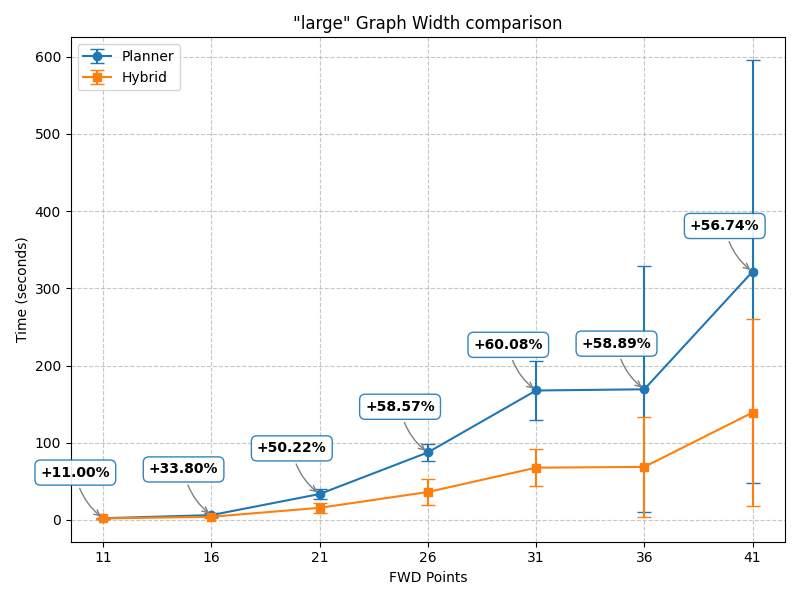}
        \caption{Same comparison for large graphs, mind the scale and the StDev.}
        \label{fig:graph_var_b}
    \end{subfigure}
    \caption{Computation speed comparison relatively to changes in the number of FWD Points for different graph sizes. The data shown is an average of 10 trips between random locations across the globe on \texttt{short} distances.}
    \label{fig:graph_var}
\end{figure}

These different cases were put to the test by simulating random trajectories for each value of width and different values of LAT (just like with FWD Points). The results can be consulted in Figure \ref{fig:graph_var_a} and Figure \ref{fig:graph_var_b}.
The first thing to note is that the overall contribution of the approach did not change: apart from very small graphs, the hybrid method's computation times are around half what the standalone planner can achieve, touching up to -60\% in large graphs. Therefore, the lateral width of the graph does not influence the feasibility of the method. What is actually different between the graphs is the nature of the task itself: not only the overall computation times increase from small to large networks (even tenfold, for 41 FWD Points) but the the standard deviation of the plans too, almost reaching the solving value itself. For this reason, individual data can be less significant, but the average is still very promising.

Lastly, the graph can be significantly altered using the \textbf{LAT Points} parameter. Increasing (or decreasing) the possible lateral routes w.r.t. the default value of 11 can significantly impact the performance of our model, especially since we've set it up as to cut the search region to a variable percentage of the input size. 

\begin{table}[h]
    \centering
    \begin{tabular}{|c|c|c|c|c|c|}
        \hline
        \textbf{LAT} & \textbf{Planner Time} & \textbf{Hybrid Time} & \textbf{\% diff.} & \textbf{Planner Fuel} & \textbf{Hybrid Fuel} \\
        \textbf{Points} & (average) & (average) & & (average) & (average) \\
        \hline
        5 & 3.34 s & 2.45 s & \textbf{-26.56\%} & 952.98 kg & 952.98 kg \\
        11 & 13.70 s & 6.33 s & \textbf{-53.80\%} & 1101.65 kg & 1101.65 kg \\
        15 & 14.38 s & 11.22 s & \textbf{-21.98\%} & \textbf{975.84 kg} & 977.26 kg \\
        21 & 19.63 s & 21.17 s & +7.85\% & \textbf{980.88} & 983.26 \\
        25 & 16.78 s & 17.53 s & +4.47\% & \textbf{1005.67} & 1010.15 \\
        31 & 21.95 s & 22.55 s & +2.73\% & \textbf{1003.17} & 1006.83 \\
        \hline
    \end{tabular}
    \caption{Comparison of computation speed between standalone planner and hybrid model on graph width set to \texttt{large} for every considered value of FWD Points. All simulated trajectories are \texttt{short} trips calculated on 21 FWD Points at \texttt{normal} width.}
    \label{tab:lat_p}
\end{table}

\newpage

As it's evident by the results in Table \ref{tab:lat_p}, while our model works best with the default value (as is to be expected, since it was used to fine tune both the RL Agent and the pipeline itself), increasing the width density of the graph lowers the improvement of our method, up to a breaking point where it's suboptimal altogether. This could be attributed to both the parametric approach of the region's width and the limited number of moves of the solver. Since we've established that we wanted to remain within 1/3 of the region's original size while cutting the graph, increasing the total number of lateral trajectories also increases the final number of options. However, as established in Section \ref{sec:results}, decreasing this limit further could lead to fuel sub-optimality, and that's already the case in minimal part with the trips shown. Furthermore, since the solver at every step can only visit up to 3 lateral locations (see Section \ref{sec:hybrid_model}), constantly increasing the LAT Points parameter without also lengthening the single trajectory (i.e. the number of FWD Points) means that increasingly more points will not contribute to the solving process; it follows that removing a point not used to begin with does not increase computation speed.

%% file: Appendix.tex
\appendix
\section{Raw data}

\setlength{\belowcaptionskip}{10pt}

\begin{table}[h]
    \centering
    \begin{tabular}{|c|c|c|c|c|c|c|}
        \hline
        \textbf{FWD points} & \textbf{Solver} & \textbf{Std. Dev.} & \textbf{Hybrid} & \textbf{Std. Dev} & \textbf{\% diff.} \\
         & \textbf{(average)} & & \textbf{(average)} & & \\
        \hline
        11 & 1.82s & 0.49 & 3.02s (1.59) & 0.57 (0.09) & +65.93\% \\
        16 & 4.25s & 0.99 & 5.16s (1.52) & 0.84 (0.07)& +21.41\%\\
        21 & 13.70s & 3.68 & 6.96s (1.63) & 2.02 (0.08) & \textbf{-49.20\%} \\
        26 & 35.99s & 15.99 & 16.83s (1.44) & 6.05 (0.07) & \textbf{-53.24\%} \\
        31 & 66.62s & 35.56 & 29.94s (1.37) & 12.79 (0.06) & \textbf{-56.06\%} \\
        36 & 75.17s & 44.31 & 34.65s (1.48) & 16.02 (0.06) & \textbf{-54.91\%} \\
        41 & 145.83s & 79.28 & 69.11s (1.29) & 31.87 (0.07) & \textbf{-53.61\%} \\
        \hline

    \end{tabular}
    \caption{Computation time for each graph size, averaged on 10 random trips between Europe airports for each row, measured with both the original implementation and the hybrid algorithm. For the Hybrid model's time, the value in parenthesis represents the amount of time consumed by the Agent (which is included in the total value). All trips are generated are of the \texttt{short} variation and are computed with a \texttt{normal} graph width.}
    \label{tab:5_trips}
\end{table}

\begin{table}[h]
    \centering
    \begin{tabular}{|c|c|c|c|c|}
        \hline
        \textbf{Fuel (Planner)} & \textbf{Fuel (Hybrid)} & \textbf{Time (Planner)} & \textbf{Time (Hybrid)} & \textbf{\% diff.} \\
        \hline
        2449.28 kg & 2449.28 kg & 393.45s & 195.73s & \textbf{-50.26\%} \\
        2490.88 kg & 2490.88 kg & 1460.07s & 584.91s & \textbf{-59.94\%} \\
        2387.16 kg & 2387.16 kg & 511.73s & 245.05s & \textbf{-52.12\%} \\
        2410.10 kg & 2410.26 kg & 958.83s & 530.23s & \textbf{-44.71\%} \\
        2430.12 kg & 2430.12 kg & 820.45s & 430.23s & \textbf{-47.57\%} \\
        \hline
    \end{tabular}
    \caption{Fuel consumption consistency and computation time comparison between hybrid method and standalone planner for \texttt{long} trips at 21 FWD points. These represent single trips as opposed to averages of multiple iterations due to their complexity to compute (keep in mind that the 'solving time' is not the whole time needed to extract data).}
    \label{tab:fuel_time_data}
\end{table}

\begin{table}[]
    \centering
    \begin{tabular}{|c|c|c|c|c|c|}
        \hline
        \multicolumn{6}{|c|}{\textbf{Thin Graph}} \\
        \hline
        \textbf{FWD Points} & \textbf{Planner Time} & \textbf{StDev} & \textbf{Hybrid Time} & \textbf{StDev} & \textbf{\% diff.} \\
        \hline
        11 & 1.08 s & 0.14 & 1.96 s (0.15) & 0.10 & +45.10\% \\
        16 & 2.75 s & 0.66 & 3.15 s (0.17) & 0.46 & + 13.70\% \\
        21 & 6.89 s & 1.73 & 3.81 s (0.18) & 0.52 & \textbf{-44.66\%} \\
        26 & 10.90 s & 4.29 & 5.88 s (0.17) & 1.57 & \textbf{-46.03\%} \\
        31 & 23.10 s & 7.35 & 11.45 s (0.18) & 4.48 & \textbf{-50.41\%} \\
        36 & 32.48 s & 14.25 & 16.74 s (0.19) & 8.71 & \textbf{-48.47\%} \\
        41 & 46.66 s & 20.42 & 22.84 s (0.17) & 10.12 & \textbf{-51.05\%} \\
        \hline
    \end{tabular}
    \caption{Comparison of computation speed between standalone planner and hybrid model on graph width set to \texttt{small} for every considered value of FWD Points.}
    \label{tab:small_graph}
\end{table}

\begin{table}[]
    \centering
    \begin{tabular}{|c|c|c|c|c|c|}
        \hline
        \multicolumn{6}{|c|}{\textbf{Large Graph}} \\
        \hline
        \textbf{FWD Points} & \textbf{Planner Time} & \textbf{StDev} & \textbf{Hybrid Time} & \textbf{StDev} & \textbf{\% diff.} \\
        \hline
        11 & 2.26 s & 0.59 & 2.01 s (0.19) & 0.46 & \textbf{-11.00\%} \\
        16 & 6.32 s & 2.71 & 4.18 s (0.19) & 1.90 & \textbf{-33.80\%} \\
        21 & 33.62 s & 6.64 & 15.74 s (0.18) & 6.74 & \textbf{-50.22\%} \\
        26 & 87.50 s & 10.51 & 36.25 s (0.19) & 16.05 & \textbf{-58.57\%} \\
        31 & 167.74 s & 37.88 & 67.75 s (0.20) & 23.92 & \textbf{-60.08\%} \\
        36 & 169.27 s & 159.53 & 68.77 s (0.18) & 64.87 & \textbf{-58.89\%} \\
        41 & 321.68 s & 273.80 & 139.14 s (0.19) & 120.70 & \textbf{-56.74\%} \\
        \hline
    \end{tabular}
    \caption{Comparison of computation speed between standalone planner and hybrid model on graph width set to \texttt{large} for every considered value of FWD Points.}
    \label{tab:large_graph}
\end{table}
